\documentclass[letterpaper]{article} 
\usepackage{aaai25}  
\usepackage{times}  
\usepackage{helvet}  
\usepackage{courier}  
\usepackage[hyphens]{url}  
\usepackage{graphicx} 
\usepackage{natbib}  
\usepackage{caption} 
\usepackage{algorithm}
\usepackage{algorithmic}

\usepackage{amsmath}
\usepackage{amssymb}
\usepackage{cuted}
\usepackage{xcolor}

\usepackage{newfloat}
\usepackage{listings}
\DeclareCaptionStyle{ruled}{labelfont=normalfont,labelsep=colon,strut=off} 
\floatstyle{ruled}
\newfloat{listing}{tb}{lst}{}
\floatname{listing}{Listing}
\title{Toy-GS: Assembling Local Gaussians for Precisely Rendering Large-Scale 
Free Camera Trajectories}
\author{
    Xiaohan Zhang\textsuperscript{\rm 1},
    Zhenyu Sun\textsuperscript{\rm 1},
    Yukui Qiu\textsuperscript{\rm 1},
    Junyan Su\textsuperscript{\rm 1},
    Qi Liu\textsuperscript{\rm 1}
    \thanks{Corresponding author}
}
\affiliations{
    \textsuperscript{\rm 1}South China University of Technology, Guangzhou, China\\

    \{ftxiaohanzhang, 202264690427, 202162311356, ft\_su.junyan\}@mail.scut.edu.cn\\ 
    drliuqi@scut.edu.cn
}

\usepackage{bibentry}

\begin{document}

\maketitle

\begin{figure*}[t]
\centering
\includegraphics[width=1\textwidth]{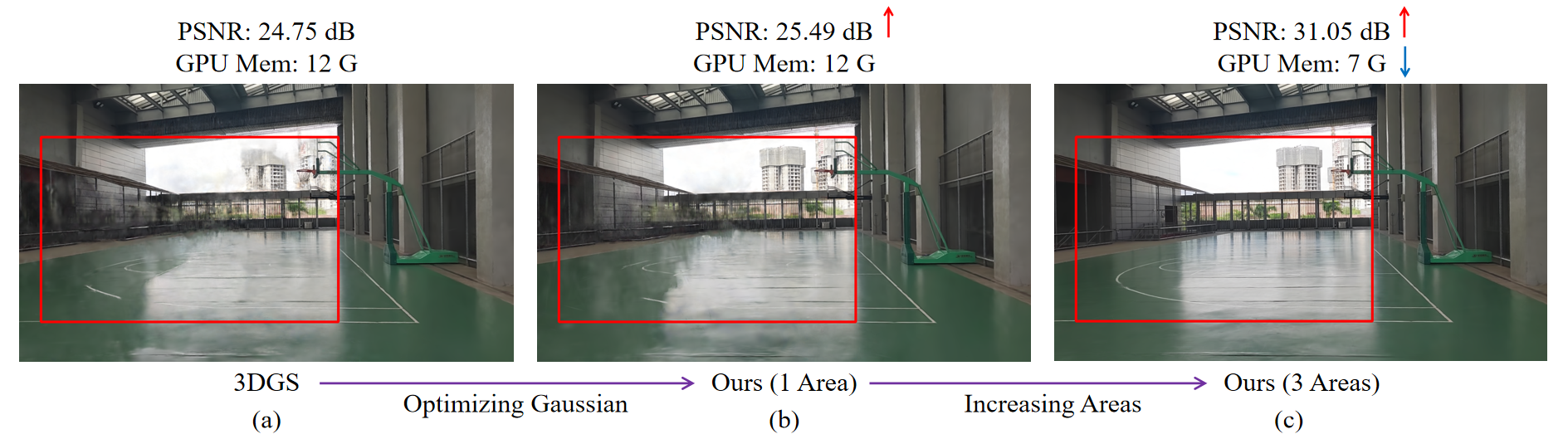} 
\caption{Toy-GS improves rendering quality while reducing GPU memory consumption. Utilizing a scene in datasets for example: (a) The rendering result of the original Gaussian Splatting. (b) By incorporating the multi-view constraint and position-aware point adaptive control into 3DGS, we enhance the background and texture details' rendering precision, resulting in the improvement of 0.74 dB in PSNR. (c) We improve by 5.56 dB in PSNR while reducing 5 G of GPU memory by dividing the entire scene into three areas and training a local Gaussian for each region separately. We design a local-global rendering method that can fully utilize the texture information of the local Gaussians to enhance the rendering effect.
} 
\label{intro_con}
\end{figure*}

\begin{abstract}
Currently, 3D rendering for large-scale free camera trajectories, namely, arbitrary input camera trajectories, poses significant challenges: 1) The distribution and observation angles of the cameras are irregular, and various types of scenes are included in the free trajectories; 2) Processing the entire point cloud and all images at once for large-scale scenes requires a substantial amount of GPU memory. This paper presents a Toy-GS method for accurately rendering large-scale free camera trajectories. Specifically, we propose an adaptive spatial division approach for free trajectories to divide cameras and the sparse point cloud of the entire scene into various regions according to camera poses. Training each local Gaussian in parallel for each area enables us to concentrate on texture details and minimize GPU memory usage. Next, we use the multi-view constraint and position-aware point adaptive control (PPAC) to improve the rendering quality of texture details. In addition, our regional fusion approach combines local and global Gaussians to enhance rendering quality with an increasing number of divided areas. Extensive experiments have been carried out to confirm the effectiveness and efficiency of Toy-GS, leading to state-of-the-art results on two public large-scale datasets as well as our SCUTic dataset. Our proposal demonstrates an enhancement of 1.19 dB in PSNR and conserves 7 G of GPU memory when compared to various benchmarks.
\end{abstract}

%
\begin{links}
    \link{Code}{https://drliuqi.github.io/}
\end{links}

\section{Introduction}

\begin{figure*}[t]
\centering
\includegraphics[width=1\textwidth]{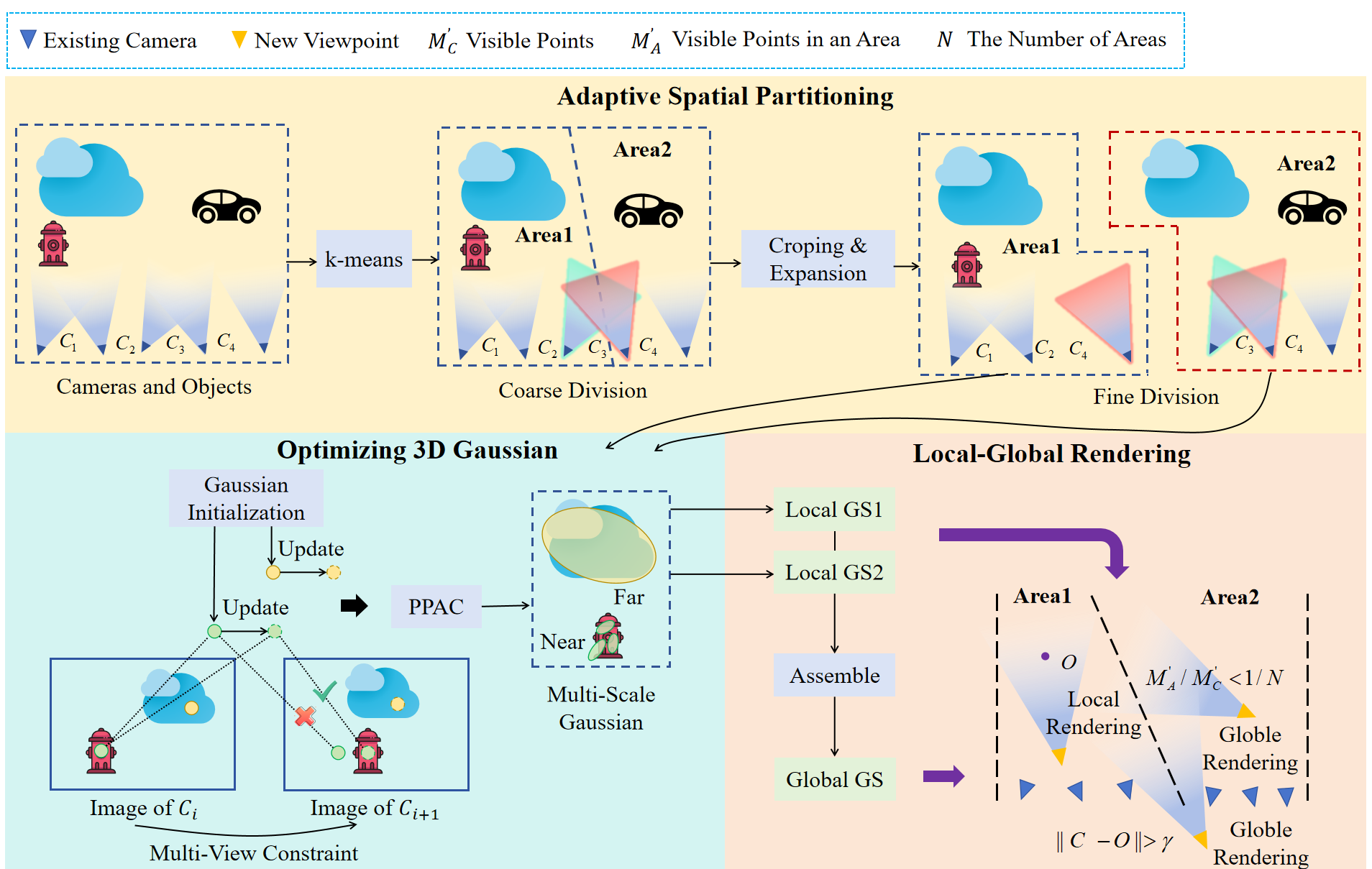} 
\caption{The pipeline of Toy-GS. Firstly, we adaptively divide cameras and the point cloud into multiple areas based on the camera poses to align the point cloud distribution with the distribution of camera poses in each region. Next, we use the multi-view constraint to enhance texture details' rendering and the position-aware point adaptive control to improve distant objects' rendering, leading to better overall results. Finally, we develop the local-global rendering that utilizes both local and global Gaussians effectively to improve rendering accuracy and reduce GPU memory consumption as the number of areas increases. 
} 
\label{pipeline}
\end{figure*}

Large-scale free camera trajectory refers to the long and irregular camera movement during photography, which contains multiple objects of interest \cite{wang2023f2}. Two characteristics of large-scale free trajectories can result in decreased performance when creating new viewpoints. On the one hand, spatial representation capacity is not evenly distributed. Specifically, when the path is long and narrow, many parts of the scenes are left empty and cannot be seen from any viewing angle. Foreground objects in the visible regions are captured with dense and close input views, while sparse and distant input views cover the background areas. On the other hand, large-scale scenes consist of a high quantity of images, requiring a significant amount of GPU memory. Furthermore, these datasets include a wide range of objects and intricate texture details, adding complexity to the rendering process. 

From two viewpoints, F2-NeRF \cite{wang2023f2} proposed how to perform high-quality viewpoint synthesis along extensive free trajectories with high quality: 1) Adaptive spatial division divides the entire scene into multiple areas, allowing cameras in each area to focus on finer texture details and enhance rendering accuracy; 2) Warping unbounded scenes to bounded ones enhances the rendering effect of distant objects. 

Recently, 3D Gaussian Splatting (3DGS) \cite{kerbl20233d} has surpassed neural radiance fields (NeRF) \cite{mildenhall2021nerf} in viewpoint rendering, achieving a new state-of-the-art result. 3DGS takes into account not just camera poses, but also a spatial sparse point cloud, providing additional structural priors to improve rendering accuracy beyond what NeRF offers. Nevertheless, 3DGS necessitates additional GPU memory. This method accomplishes the projection of all Gaussian ellipsoids onto a viewpoint's imaging plane simultaneously to compute the rendered image, while NeRF creates a ray for each pixel to batch-train the entire image. The larger the scene, the more GPU memory is consumed, which is not ideal for reconstructing large-scale scenes. To solve this problem, VastGaussian \cite{lin2024vastgaussian} suggested partitioning the space into several regions, training a separate Gaussian model for each region, and then combining all the local Gaussian models into one global Gaussian model. VastGaussian excelled in processing large aerial datasets \cite{turki2022mega}, but the rendering accuracy decreased when rendering large-scale free trajectories on the ground with imbalanced information. Several factors influence the outcome of the reconstruction. (1) In aerial datasets, the distribution of cameras is relatively dense. The camera trajectory for each building can be seen from various angles, resembling object-centric trajectories of $360^{\circ}$. Nevertheless, when using free trajectory datasets, the lack of varied viewpoints for objects means that relying solely on a global Gaussian in VastGaussian will result in holes in the rendered images. (2) Cameras positioned at ground level shooting upward capture expansive scenes, however, 3DGS lacks feature points in these expansive areas, leading to subpar rendering effects in those regions. (3) VastGaussian utilizes bounding boxes to divide regions. Incorrectly setting the boundary range can result in an unfair distribution of cameras in each area.

To address them, we propose the Toy-GS method for rendering large-scale free camera trajectories. Figure \ref{intro_con} demonstrates how our approach can effectively decrease GPU memory usage and enhance rendering quality as the number of regions expands. Our contributions are as follows:

\begin{itemize}
\item We propose an adaptive spatial division for large-scale free trajectories. We calculate the cameras' clustering model based on camera poses, which is then used to segment the entire point cloud into specific areas. This aligns the point cloud distribution with the distribution of camera poses in each area, enhancing rendering accuracy.
\item We use the multi-view constraint to enhance areas with noticeable differences in depth and normal, boosting the quality of texture details in the rendering. Additionally, we apply the position-aware point adaptive control (PPAC) algorithm to improve the rendering quality for distant landscapes.
\item We design a rendering method that combines information from local Gaussians on a global scale. Our method shows a better rendering effect with an increasing number of regions compared to VastGaussian's global Gaussian rendering method.
\item We build a new extensive dataset of free trajectories that is 2 to 3 times larger than existing ones, increasing the difficulty of synthesizing new perspectives.
\end{itemize}

\section{Related Work}
\textbf{Novel View Synthesis.} The Novel View Synthesis (NVS) is a process that generates new viewpoint images from given posed images. Lumigraph \cite{buehler2001unstructured,gortler2023lumigraph} and light field functions \cite{davis2012unstructured,levin2010linear} were widely used in the NVS field to directly interpolate the provided images. To enhance the quality of the generated image, a variety of methods were employed that involved an explicit 3D reconstruction of the scene using meshes \cite{debevec2023modeling,thies2019deferred,waechter2014let,wood2023surface}, voxels \cite{he2020deepvoxels++,lombardi2019neural,lombardi2021mixture}, point clouds \cite{aliev2020neural,xu2022point}, depth maps \cite{dhamo2019peeking,shade1998layered,shih20203d,tulsiani2018layer}, and multi-plane images (MPI) \cite{flynn2019deepview,li2020crowdsampling,mildenhall2019local,srinivasan2019pushing,tucker2020single,zhou2018stereo}. These 3D reconstructions were then utilized to generate new images. Lately, techniques utilizing implicit 3D representations have demonstrated impressive rendering results in view synthesis. NeRF \cite{mildenhall2021nerf} sampled along view rays to determine the color and transparency of each point, then integrated the volume to obtain the final pixel color. 3DGS employed the sparse point cloud of the scene as a reference, created a Gaussian ellipsoid at each point, and projected all ellipsoids onto the viewpoint to generate the rendered image. Implicit reconstruction is more effective than explicit reconstruction in accurately portraying information that varies depending on perspectives.

\noindent\textbf{Rendering on Free Camera Trajectories.} Existing methods have successfully attained a high level of rendering quality for both the camera's $360^{\circ}$ object-centric trajectories and forward-facing trajectories. Instant-NGP \cite{muller2022instant} used a hash grid to gather additional spatial data and a shallow MLP to achieve quick and high-quality rendering. NeRF++ \cite{zhang2020nerf++} decoupled the whole scene into bounded and unbounded scenes and rendered them individually, enhancing the rendering quality for distant objects. Mip-NeRF-360 \cite{barron2022mip} suggested utilizing view cones rather than view rays to enhance rendering accuracy by capturing additional spatial information. Rendering becomes more difficult with the free trajectory as it includes more complex camera poses and objects of interest than the previous two trajectories. F2-NeRF \cite{wang2023f2} designed a spatial partitioning method and a compression method for unbounded scenes based on camera poses. This enables it to focus on finer texture details, ultimately enhancing the rendering quality. 3DGS can achieve superior rendering results compared to NeRF. Moreover, the use of 3DGS in extensive settings is growing in popularity.

\noindent\textbf{Gaussian Splatting on Large-Scale Scenes.} 3DGS has shown excellent rendering results on large-scale aerial datasets. VastGaussian \cite{lin2024vastgaussian} proposed dividing the scene into various regions, training the Gaussians in each region concurrently, and combining the Gaussians from all regions to generate new perspectives. CityGaussian \cite{liu2024citygaussian} proposed segmenting the entire scene according to the visibility of the point cloud and developing a Level of Detail (LOD) algorithm for rendering different perspectives. In aerial datasets, there is a high density of cameras. Each building can be viewed from various perspectives, resulting in camera movements that resemble $360^{\circ}$ object-centric trajectories. Using Gaussian-based algorithms to generate free trajectories with uneven information results in a decrease in rendering quality. To this end, we propose a novel rendering method Toy-GS for large-scale free trajectories. We divide the scene dynamically according to the camera's pose, improve texture details and rendering effects of distant objects, and introduce a rendering method that combines local and global Gaussian information to enhance rendering quality and save GPU memory.

\section{Method}


Toy-GS is capable of accurately rendering large-scale free camera trajectories while decreasing GPU memory usage. The pipeline of our method is shown in Figure \ref{pipeline}. The adaptive division of large-scale free trajectories is discussed in Section 3.1. The enhancement of 3DGS rendering by the multi-view constraint and PPAC is shown in Section 3.2. The enhancement of rendering effects by local-global fusion is discussed in Section 3.3.

\begin{figure}[t]
\centering
\includegraphics[width=1\columnwidth]{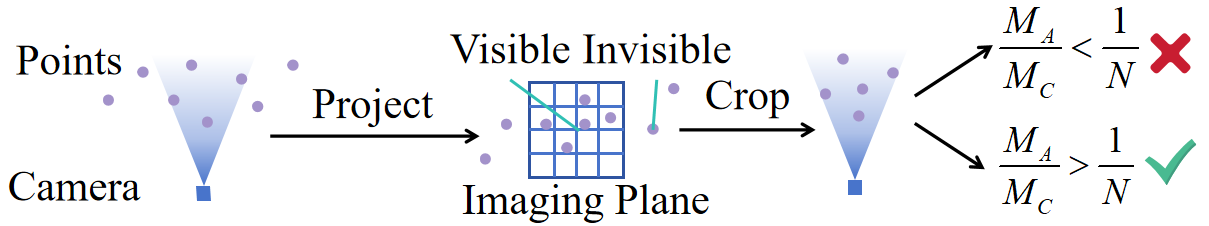} 
\caption{Adaptive camera selection strategy in an area. For each camera, we project all points onto the imaging plane and filter out points that are not visible to this camera. We then determine if there are enough points visible to the camera within this area to decide whether to select this camera.}
\label{points_visible}
\end{figure}

\subsection{Adaptive Spatial Partitioning}
Based on the camera poses, we segment the cameras and the point cloud into different regions. Most cameras in each area can capture the point cloud of that specific region, which enhances the accuracy of the rendering. Initially, we train a k-means model using the camera poses, which helps divide the camera and point cloud into several areas $\{A_n\}_{n=1}^N$. To fully train the local Gaussians, we then need to crop and expand cameras in each region. Specifically, we choose a particular area to train the local Gaussian, exclude cameras in this area that cannot see it, and add cameras from other areas that can view it. We use the visibility of points to prune and expand the camera. If a camera captures certain points in this area, it is deemed to be within the region; if not, it is considered outside of it. We calculate the points visible to each camera through projection. Using the intrinsic matrix $\mathbf{K}$ and the transformation matrix $\mathbf{T}$, the spatial point $\mathbf{P}(x, y, z)$ can be projected onto the imaging plane $\mathbf{p}(u, v)$:
\begin{equation}
z\mathbf{p}=\mathbf{K}\mathbf{T}\mathbf{P}
\end{equation}

If a camera has a viewing range of $M_c$, it can observe $M_{A_i}$ points in a specific area $A_i$, with a total of $N$ regions. As the area divisions increase, the value of $M_{A_i}$ decreases, leading us to adjust the number of cameras based on the following conditions:
\begin{equation}
mask=\left\{
\begin{array}{cl}
1, & M_{A_i}/M_c > 1/N  \\
0, & \text{else} \\
\end{array} \right.
\end{equation}
A mask value of 1 signifies that the camera has permission to view this particular area. When the mask is set to 0, it means that the camera is unable to view this area or can only capture a limited portion of the scene, leading to the removal of this camera, as shown in Figure \ref{points_visible}. Moreover, our strategy robustly partitions cameras into corresponding regions under varying point cloud densities.  Even when the point cloud density varies, a stable ratio is obtained because 
$M_{A_i}$ and $M_c$ increase or decrease simultaneously. To fully train the local Gaussian of this region, we project all observable points from cameras in this region into space to create the initial point cloud. This ensures that point clouds in each region have maximum visibility from cameras in that region, and more accurate local Gaussians can be trained.

\subsection{The Rendering Optimization}
At each point in the input point cloud, 3DGS initializes an anisotropic 3D Gaussian $G(x)$ with the notation:
\begin{equation}
G(\mathbf{x})=e^{-\frac{1}{2}(\mathbf{x}-\boldsymbol{\mu})^{T} \mathbf{\Sigma}^{-1}(\mathbf{x}-\boldsymbol{\mu})}    
\end{equation}
where $\mu\in\mathbf{R}^{3\times1}$ and $\mathbf{\Sigma}\in\mathbf{R}^{3\times3}$ denote the mean vector and covariance matrix, respectively. Continuously optimizing the shape attributes and color features of the Gaussian ellipsoids and rendering the pixel $\mathbf{p}$ through integral by $N$ ordered Gaussians $\{G_i|i=1,..., N\}$ overlapping $\mathbf{p}$: 
\begin{equation}
c(p) = \sum_{i=1}^{N} c_i \alpha_i \prod_{j=1}^{i-1} (1 - \alpha_j)
\end{equation}
where $c_i$ and $a_i$ represent the color and opacity of Gaussian $G_i$.
However, because depth information is lacking, the positions $\mu$ of many Gaussians are incorrect, resulting in a decrease in rendering quality. Additionally, the rendering quality of distant objects like clouds is diminished because Gaussian points are restricted to a specific area for distribution.

To address both issues, we implement the Patchmatch algorithm \cite{barnes2009patchmatch} to guide the Gaussians to their proper locations using multi-view constraints, thereby improving the texture detail rendering effect. To enhance the rendering effect of distant objects, we utilize PPAC method \cite{chen2023periodic}  to apply Gaussians of different scales to the foreground and background. 

First, we render the depth map and normal map for each image. The depth maps are determined by weighted blending of overlapping Gaussians based on their opacity and spatial distribution. The normal maps are computed through weighted blending of contributing Gaussians based on their opacity and normals. Then we use Patchmatch to move the Gaussian ellipsoids to their correct positions via multi-view constraints. Patchmatch sets up a spatial plane $(d,\mathbf{n})$ for every pixel $\mathbf{p}$ in the reference view, with $\mathbf{n}$ representing the pixel's rendered normal and $d$ indicating the distance from the camera coordinate origin to the spatial plane.
Assuming accurate values for the normal $\mathbf{n}$ and depth $d$, the pixel $\mathbf{p}$ can be projected to its corresponding pixel $\mathbf{p}'$ in the source view using homography warping:
\begin{equation}
\mathbf{p}'=\mathbf{H}\mathbf{p}
\end{equation}
where $\mathbf{H}$ is denoted as:
\begin{equation}
\mathbf{H}=\mathbf{K}\left(\mathbf{R}-\frac{\mathbf{t} \mathbf{n}^{\top}}{d}\right) \mathbf{K}^{-1}
\end{equation}
in which $\mathbf{R}$ and $\mathbf{t}$ represent the rotation and translation of the camera coordinate system for reference view and source view, respectively, and $\mathbf{K}$ is the camera’s intrinsic matrix. Thus, minimizing the discrepancy between $\mathbf{H}\mathbf{p}$ and $\mathbf{p}'$ can enhance the accuracy of the depth map produced by 3DGS.
Gaussian ellipsoids are initialized at positions with incorrect depth estimation by comparing the initial depth map with the optimized depth map. This method improves the geometric structure of 3DGS and boosts the rendering quality of texture details.
PPAC advocates utilizing larger points for distant locations and smaller points for nearby areas. The scale factor $\gamma(\mathbf{\mu})$ of each Gaussian location $\mathbf{\mu}$ is defined as:
\begin{equation}
\gamma(\boldsymbol{\mu})=\left\{\begin{array}{lrl}1 & \text { if } & \|\boldsymbol{\mu}\|_{2}<2 r \\\|\boldsymbol{\mu}\|_{2} / r-1 & \text { if } & \|\boldsymbol{\mu}\|_{2} \geq 2 r\end{array}\right.
\end{equation}
where $r$ denotes the scene radius. By using large Gaussians to cover distant scenes, 3DGS can effectively render distant scenes and enhance the quality of faraway objects. 

\subsection{Local-Global Rendering}

\begin{figure}[t]
\centering
\includegraphics[width=1\columnwidth]{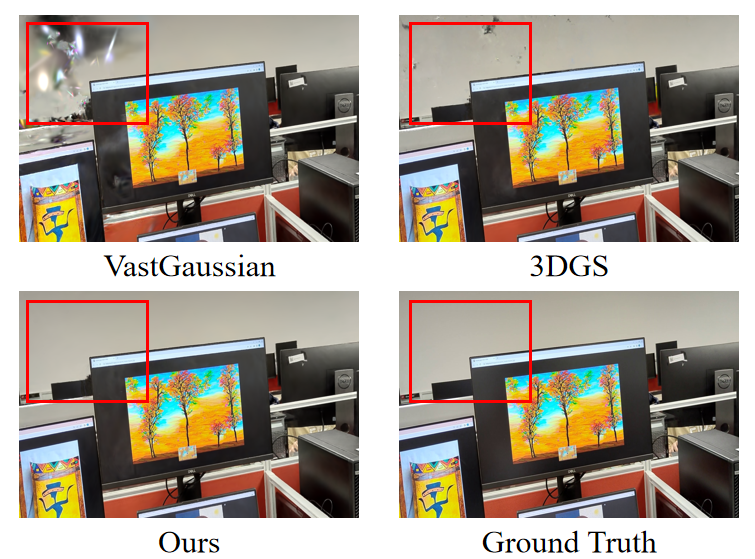} 
\caption{Comparison of different rendering methods. VastGaussian prunes each region's 3DGS, leading to holes in the rendering process. When there are a few viewpoints, training the entire scene with 3DGS outperforms VastGaussian. Our approach maximizes the use of local Gaussians to enhance rendering quality.
}
\label{render_opti}
\end{figure}

Splitting the scene into different sections and training a 3DGS in each one can help decrease GPU memory usage. While rendering, VastGaussian suggests eliminating the Gaussian ellipsoids located outside each area and combining all the trimmed local 3DGSs into a global 3DGS to generate new perspectives, reducing the number of floaters. Pruning the 3DGS in aerial datasets has minimal impact on the rendering effect because of the dense camera viewpoints. However, since free trajectory datasets have limited camera viewpoints on each object, pruning the 3DGS may result in holes in images that diminish the rendering quality. To address this issue, we design a local-global rendering method that fully utilizes the information from the local Gaussians to improve rendering quality as illustrated in Figure \ref{render_opti}. In Section 3.1, we prune and expand cameras and points in each area, allowing for comprehensive training of Gaussian to render viewpoints that can observe this area. However, it is unreasonable to use a specific Gaussian to represent this viewpoint when a new camera can view multiple regions. Hence, we create the adaptive criteria to assess if a camera can be depicted by a particular local Gaussian. In the rendering phase, when we have a new viewpoint $\mathbf{C}_{new}$, we initially identify the corresponding area $A_i$ by applying the k-means model outlined in Section 3.1. Next, we determine the distance from the center $\mathbf{O}$ to every current camera in the area, selecting the greatest distance as the threshold $\gamma$. The distance constraint is defined as:
\begin{equation}
\|\mathbf{C}_{new}-\mathbf{O}\|<\gamma
\end{equation}
Next, we calculate the number of points $M_{c_{new}}'$ visible from this viewpoint based on the trained dense point cloud and determine how many points belong to this region $M_{A_i}'$. When the number of regions is $N$, the visibility constraint is:
\begin{equation}
M_{A_i}'/M_{c_{new}}' > 1/N
\end{equation}
If this new viewpoint meets both the distance and visibility constraints, it indicates that this new viewpoint can be rendered by this area's local Gaussian. Otherwise, this alternative perspective
can view different regions and is rendered by the global Gaussian followed by VastGaussain. 

\section{Experiments}

\begin{table*}[t]
\scalebox{0.6}{
\begin{tabular}{c|c|c|c|c|c|c|c|c|c|c|c|c|c|c|c|c}
\hline & \multicolumn{4}{|c|}{ $general$ } & \multicolumn{4}{|c|}{ $indoor$ } & \multicolumn{4}{|c|}{ $outdoor$ } & \multicolumn{4}{|c}{ $overall$ } \\
& PSNR $\uparrow$ & SSIM $\uparrow$  & LPIPS  $\downarrow$  &  mem.(M) $\downarrow$  & 
PSNR $\uparrow$  & SSIM  $\uparrow$  & LPIPS  $\downarrow$  &  mem.(M) $\downarrow$  & PSNR $\uparrow$  & SSIM  $\uparrow$  & LPIPS  $\downarrow$  &  mem.(M) $\downarrow$  & PSNR  $\uparrow$  &  SSIM $\uparrow$  & LPIPS  $\downarrow$  &  mem.(M) $\downarrow$  \\
\hline F2-NeRF & 19.53 & 0.641 & 0.411 & 17820 & 17.69 & 0.622 & 0.457 & 11362 & 20.18 & 0.679 & 0.397 & 18406 & 19.13 & 0.647 & 0.422 & 15863 \\
 3DGS & 22.55 & 0.729 & 0.319 & 17798 & 22.01 & 0.762 & 0.32 & 12242 & 23.16 & 0.72 & 0.347 & 19230 & 22.57 & 0.737 & 0.329 & 16423 \\
 VastGS (3 Areas) & 22.35 & 0.751 & 0.275 & 12596 & 21.71 & 0.753 & 0.301 & 7674 & 21.86 & 0.733 & 0.295 & 8484 & 21.97 & 0.746 & 0.29 & 9584 \\
 Ours (1 Area) & 22.84 & 0.738 & 0.309 & 18132 & 22.61 & 0.777 & 0.302 & 12436 & 23.44 & 0.728 & 0.334 & 19421 & 22.96 & 0.748 & 0.315 & 16663 \\
 Ours (3 Areas) & \textbf{23.51} & \textbf{0.767} & \textbf{0.262} & \textbf{12596} & \textbf{23.57} & \textbf{0.795} & \textbf{0.269} & \textbf{7674} & \textbf{24.25} & \textbf{0.77} & \textbf{0.266} & \textbf{8484} & \textbf{23.76} & \textbf{0.777} & \textbf{0.265} & \textbf{9584} \\
\hline
\end{tabular}
}
\caption{Quantitative comparisons with other methods in terms of PSNR, SSIM, LPIPS, and GPU memory (mem.) on our SCUTic dataset. Bold denotes the best result.}
\label{tab_ours}
\end{table*}

\begin{figure*}[t]
\centering
\includegraphics[width=0.95\textwidth]{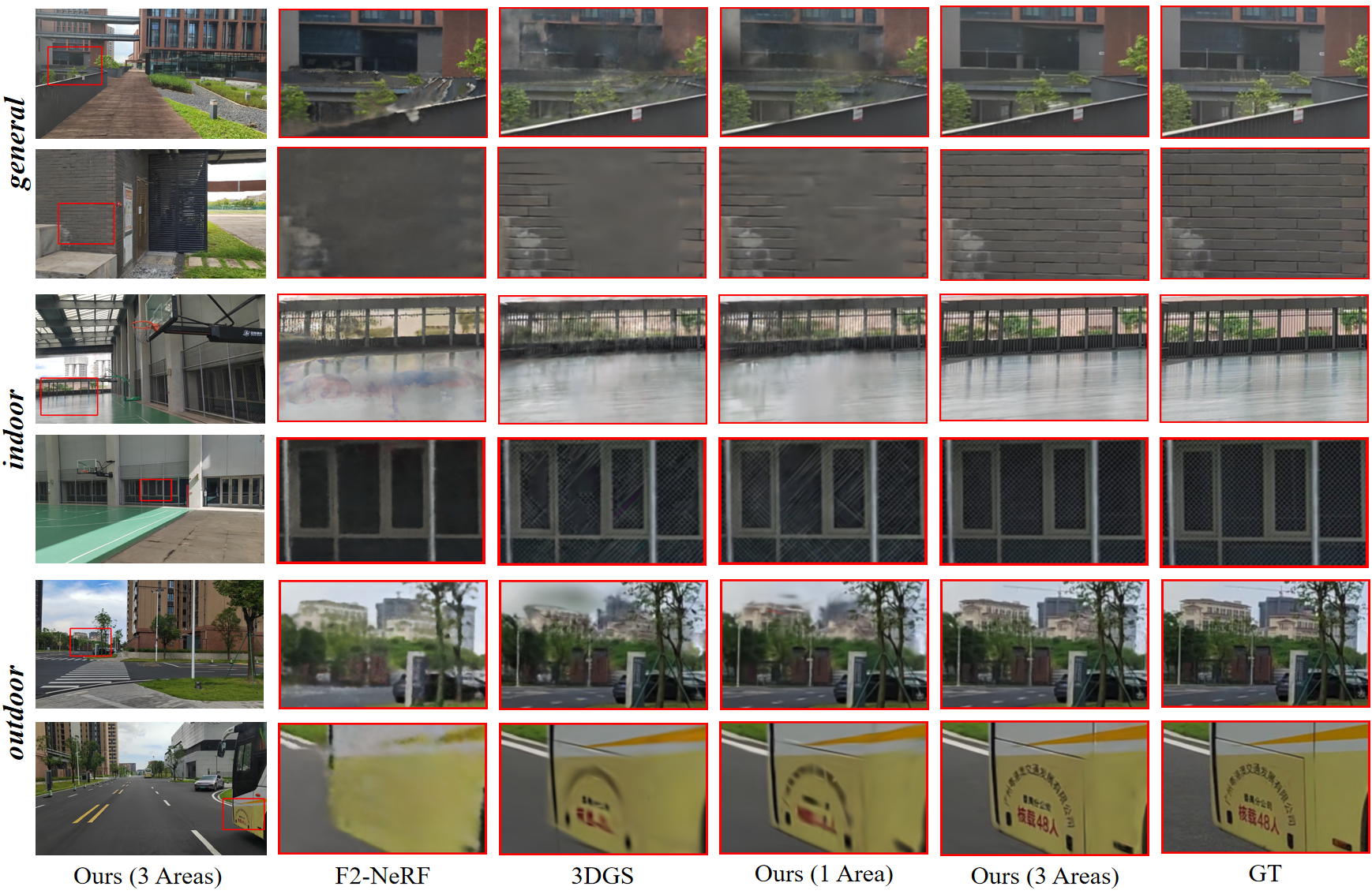} 
\caption{Visual comparisons with recent methods on our SCUTic dataset. Our method provides better rendering effects for texture details and distant objects.}
\label{ours_com}
\end{figure*}

\subsection{Datasets}
\textbf{Free dataset} \cite{wang2023f2}. F2-NeRF introduces an extensive dataset containing multiple scenes. Constructing a neural representation for the NVS task becomes highly challenging due to the narrow and elongated input camera trajectory in each scene, as well as the presence of multiple focused foreground objects. Each scene contains 200-300 images with a resolution of $1461\times935$.

\noindent\textbf{Tanks and Temples} \cite{knapitsch2017tanks}. This dataset is divided into two categories: intermediate and advanced. The intermediate category includes images taken with $360^\circ$ camera paths, while the advanced category includes images taken with free camera trajectories in both indoor and outdoor settings. We use the advanced type to evaluate the model and this dataset contains 300-400 images with a resolution of $1920\times1080$.

\noindent\textbf{SCUTic}. Our dataset, created by us, includes three main categories: $indoor$, $outdoor$, and $general$, making it a large-scale free camera dataset. The $indoor$ includes images taken along a long camera trajectory from the first floor to the third floor indoors, showcasing a diverse range of objects. The $outdoor$ includes images captured along free-motion trajectories in unbounded outdoor scenes.
The $general$ includes images captured along camera trajectories that interweave multiple times between indoor and outdoor environments, making it more challenging for viewpoints rendering. Each category includes more than 600 images with a resolution of $1920\times1080$, which can better validate the robustness of models.

\begin{table}[t]
\scalebox{0.9}{
\begin{tabular}{|c|c|c|c|c|}
\hline & PSNR $\uparrow$ & SSIM $\uparrow$ & LPIPS  $\downarrow$ & mem.(M) $\downarrow$ \\
\hline F2-NeRF  & 18.05 & 0.757 & 0.281 & 10696\\
3DGS  & 20.87 & 0.772 & 0.258 & 8372\\
VastGS & 20.11 & 0.768 & 0.246 & 7444\\ 
Ours (1 Area) & 21.12 & 0.783 & 0.243 & 8522\\
Ours (3 Areas) & \textbf{21.27} & \textbf{0.786} & \textbf{0.231} & \textbf{7444} \\
\hline
\end{tabular}
}
\caption{Quantitative comparisons with other methods in terms of PSNR, SSIM, LPIPS, and GPU memory (mem.) on the Tanks and Temples dataset. Bold denotes the best result.}
\label{tab_tanks}
\end{table}

\begin{figure}[t]
\centering
\includegraphics[width=1\columnwidth]{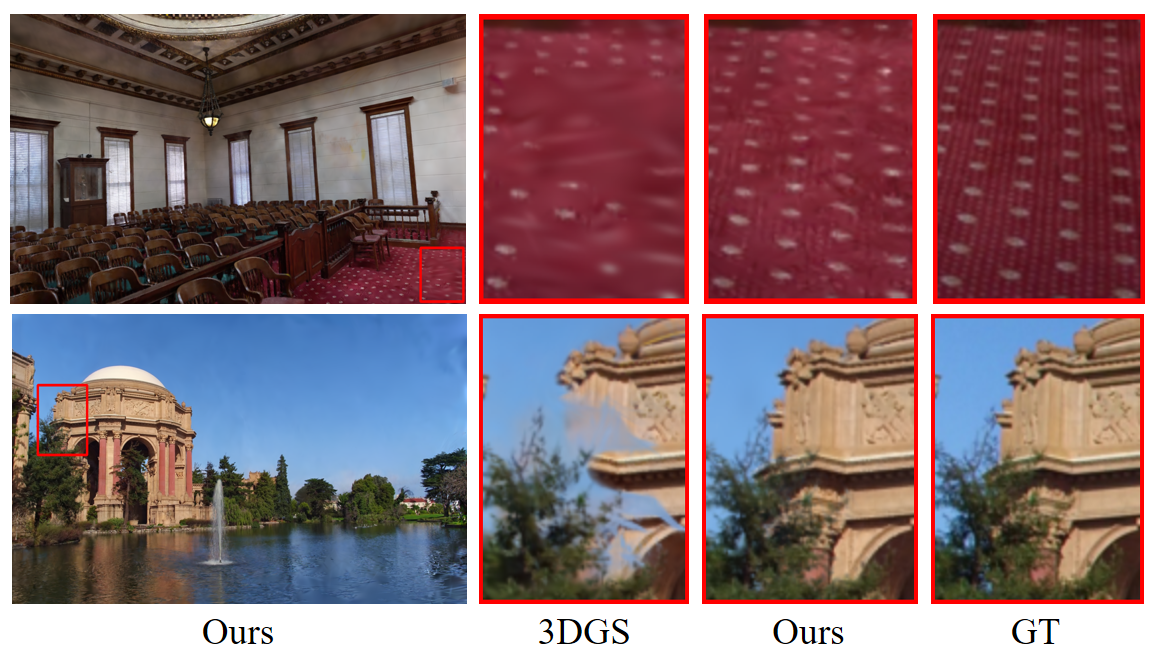} 
\caption{Visual comparisons with recent methods on the Tanks and Temples dataset. Our method produces more comprehensive objects.}
\label{tanks_com}
\end{figure}

\subsection{Results}

\begin{table}[t]
\begin{center}
\scalebox{0.9}{
\begin{tabular}{|c|c|c|c|}
\hline & PSNR $\uparrow$ & SSIM $\uparrow$  & LPIPS  $\downarrow$  \\
\hline NeRF++ & 23.47 & 0.603 & 0.499 \\
mip-NeRF-360 & 27.01 & 0.766 & 0.295 \\
Plenoxels & 22.04 & 0.537 & 0.586 \\
DVGO & 23.90 & 0.651 & 0.455 \\
InstantNGP & 24.43 & 0.677 & 0.413 \\
F2-NeRF & 26.32 & 0.779 & 0.276 \\
3DGS & 27.92 & 0.886 & 0.129 \\
VastGS & 27.90 & 0.896 & 0.104 \\ 
Ours (1 Area) & 28.83 & 0.904 & 0.104 \\
Ours (3 Areas) & \textbf{29.27} & \textbf{0.910} & \textbf{0.092} \\
\hline
\end{tabular}
}
\end{center}
\caption{Quantitative comparisons with other methods in terms of PSNR, SSIM, and LPIPS on the  Free dataset. Bold denotes the best result.}
\label{tab_free}
\end{table}


\textbf{Results on SCUTic Dataset}.
We report quantitative performance using PSNR (higher is better), SSIM (higher is better), LPIPS \cite{zhang2018unreasonable} (lower is better), as well as GPU memory (lower is better). Table \ref{tab_ours} shows the results on our SCUTic dataset. Our method ranks highest in all metrics across all scenes. We enhance the rendering effects of 3DGS using the Patchmatch and PPAC algorithms. Training our optimized Gaussian on the entire scene resulted in improvements of 0.29 dB, 0.6 dB, and 0.28 dB in PSNR on the three datasets compared to 3DGS, respectively. We adaptively partition the entire scene into three areas, train a local Gaussian in each region, and design a local-global rendering strategy to enhance rendering effects while saving GPU memory. Our method can save 5 G, 4.5 G, and 11 G of GPU memory on the three datasets by partitioning the scene into three areas, in comparison to 3DGS. Our method outperforms VastGaussian, which relies solely on the global Gaussian for generating new viewpoints, by achieving PSNR improvements of 1.16 dB, 1.8 dB, and 2.39 dB on the three datasets, respectively. Figure \ref{ours_com} displays the visual comparisons. Our approach offers distinct rendering effects for texture details as well as distant objects.

\noindent\textbf{Results on Tanks and Temples Dataset}.
The quantitative comparisons with other methods on the Tanks and Temples dataset are shown in Table \ref{tab_tanks}. Due to the relatively small scale of this dataset, our method saves 1 G of GPU memory compared to 3DGS and 3 G of GPU memory compared to F2-NeRF, while improving 0.4 dB compared to 3DGS and 3.22 dB compared to F2-NeRF. Figure \ref{tanks_com} displays visual comparisons, demonstrating that our rendered objects are more fully depicted in indoor and outdoor scenes than those produced by other methods.

\noindent\textbf{Results on Free Dataset}. We compared our results with those of several competitors using the Free dataset, which is detailed in Table \ref{tab_free}. Compared to 3DGS, our approach enhances the accuracy of rendering texture details and increases PSNR by 1.35 dB. Creating smaller sections within the scene can help to capture additional details and nuances. Splitting the scene into three areas results in a 0.44 dB improvement in PSNR compared to training the Gaussian on the entire scene.

\subsection{Ablation Studies}

\begin{table}[t]
\scalebox{0.7}{
\begin{tabular}{|c|c c c c c c c c|}
\hline & $a_{num}$ & $2P$ & $R_{G}$ & $R_{LG}$ & PSNR $\uparrow$ & SSIM $\uparrow$ & LPIPS  $\downarrow$ & mem.(M) $\downarrow$ \\
\hline (a) & 1 & & & & 22.01 & 0.762 & 0.32 & 12108 \\
(b) & 1 & $\checkmark$ & & & 22.61 & 0.777 & 0.302 & 12242 \\
(c) & 3 & $\checkmark$ & $\checkmark$ & & 21.71 & 0.753 & 0.301 & 7674 \\
(d) & 3 & $\checkmark$ & & $\checkmark$ & 23.51 & 0.795 & 0.269 & 7674 \\
(e) & 5 & $\checkmark$ & $\checkmark$ & & 21.94 & 0.761 & 0.295 & 7858 \\
(f) & 5 & $\checkmark$ & & $\checkmark$ & 23.57 & 0.795 & 0.264 & 7850 \\
(g) & 7 & $\checkmark$ & $\checkmark$ & & 22.09 & 0.768 & 0.291 & 6560 \\
(h) & 7 & $\checkmark$ & & $\checkmark$ & \textbf{23.75} & \textbf{0.797} & \textbf{0.260} & \textbf{6560} \\
\hline
\end{tabular}
}
\caption{Ablation studies of different components. $a_{num}$ is the number of areas. $2P$ is Patchmatch \& PPAC. $R_G$ is the global rendering followed by VastGaussian. $R_{LG}$ is our proposed local-global rendering method.}
\label{table_abla}
\end{table}

\begin{figure}[t]
\centering
\includegraphics[width=1\columnwidth]{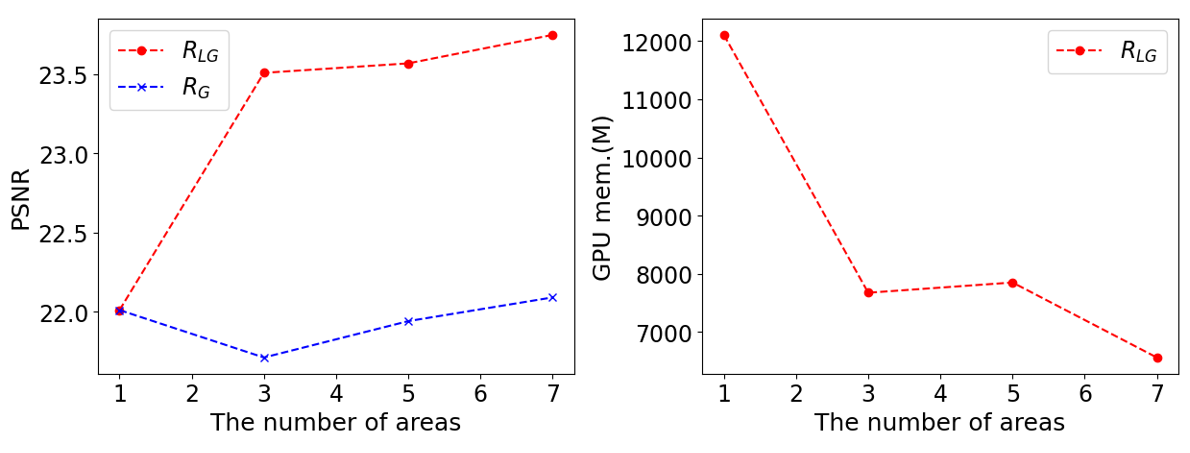} 
\caption{ Left: The PSNR of the global rendering and local-global rendering when selecting different numbers of areas. Right: The GPU memory of our method when selecting different numbers of areas.}
\label{ablation}
\end{figure}

We conduct ablations on several components to understand how these main modules work, including the number of areas $a_{num}$, Pacthmatch \& PPAC ($2P$), global rendering $R_G$, and local-global rendering $R_{LG}$. The results are shown in Table \ref{table_abla}. First, as shown in Table 4 (a) and (b), Patchmatch \& PPAC can improve the rendering accuracy of texture details, with improvements of 0.6 dB in PSNR, 0.015 in SSIM, and 0.002 in LPIPS. Second, as shown in Table 4 (d) and (b), dividing the whole scene into multiple areas can capture more texture information and save GPU memory. Our local-global rendering algorithm can fully utilize local Gaussians to enhance the rendering effect, with improvements of 0.9 dB in PSNR, 0.018 dB in SSIM, and 0.033 in LPIPS while reducing about 4.5 G GPU memory.
Third, as shown in Table 4 (d) and (c), local-global rendering improves by 1.8 dB, 0.42 dB, and 0.032 dB in PSNR, SSIM, and LPIPS, respectively. 
Fourth, as shown in Table 4 (b), (c), (e), and (g), only using the global Gaussian to render large-scale free camera trajectories with uneven information reduces the rendering quality.
The rendering effect of 7 areas decreases by 0.52 dB in PSNR, 0.009 dB in SSIM, and improves by 0.011 dB in LPIPS compared to a single area.
Fifth, as shown in Table 4 (b), (d), (f), and (h), as the number of areas increases, the rendering effect gradually improves, and the GPU memory shows a decreasing trend. Dividing into 7 regions improves by 1.14 dB, 0.02 dB, and 0.042 dB in PSNR, SSIM, and LPIPS, respectively, while reducing about 6 G GPU memory compared to a single area. Figure \ref{ablation} illustrates the correlation between the number of regions, PSNR, and GPU memory. Toy-GS can fully leverage local Gaussians to capture more texture information. Our method improves rendering quality and reduces memory usage significantly compared to the previous approaches.

\section{Conclusion}

In this paper, we present the Toy-GS, which improves rendering quality and decreases GPU memory usage for large-scale free camera trajectories by adaptively assembling local Gaussians. To begin with, we dynamically divide the cameras and point cloud of the entire scene into multiple sections according to the camera’s motion trajectory, and then train a distinct Gaussian model for each area. This method allows for capturing additional texture details while also decreasing the GPU memory usage. We improve the rendering of texture details and distant objects by using Patchmatch and PPAC. Moreover, we propose the local-global rendering algorithm that fully utilizes local Gaussian information to enhance rendering quality.
Lastly, we have developed the SCUTic dataset, consisting of free camera trajectories on a large scale, which is 2 to 3 times bigger than existing datasets. Our method achieves new state-of-the-art performance on two public datasets and our own dataset, while also cutting GPU memory usage by half.

\section{Acknowledgments}
This work was supported in part by the National Natural Science Foundation of China under Grant 62202174, in part by the Basic and Applied Basic Research Foundation of Guangzhou under Grant 2023A04J1674, and in part by The Taihu Lake Innocation Fund for the School of Future Technology of South China University of Technology under Grant 2024B105611004.

\bibliography{aaai25}

\clearpage

\thispagestyle{empty}

\section{Appendix}

\subsection{Implementation Details}
The Gaussian training involves learning rates at the start for the position, features, opacity, scale, and rotation of the Gaussian ellipsoids, which are 0.00016, 0.0025, 0.05, 0.005, and 0.001, respectively. We use Patchmatch to optimize the position of Gaussian ellipsoids between 1k and 6k iterations, with a total of 30k iterations needed. In the Patchmatch process, we execute a propagation optimization every 50 iterations, with a patch size of 20. In PPAC, the value of $r$ is set to 0.005. We apply the Adam optimizer to train the Gaussian parameters, and the experiments are deployed on a computer with a single RTX 3090 GPU.

\begin{figure}[t]
\centering
\includegraphics[width=1\columnwidth]{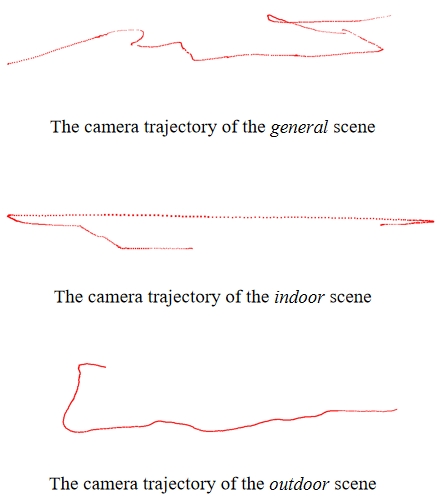} 
\caption{The camera trajectories in our SCUTic dataset. These camera trajectories are irregular and the spatial representation is not evenly distributed.
}
\label{camera_track}
\end{figure}

\begin{figure}[t]
\centering
\includegraphics[width=1\columnwidth]{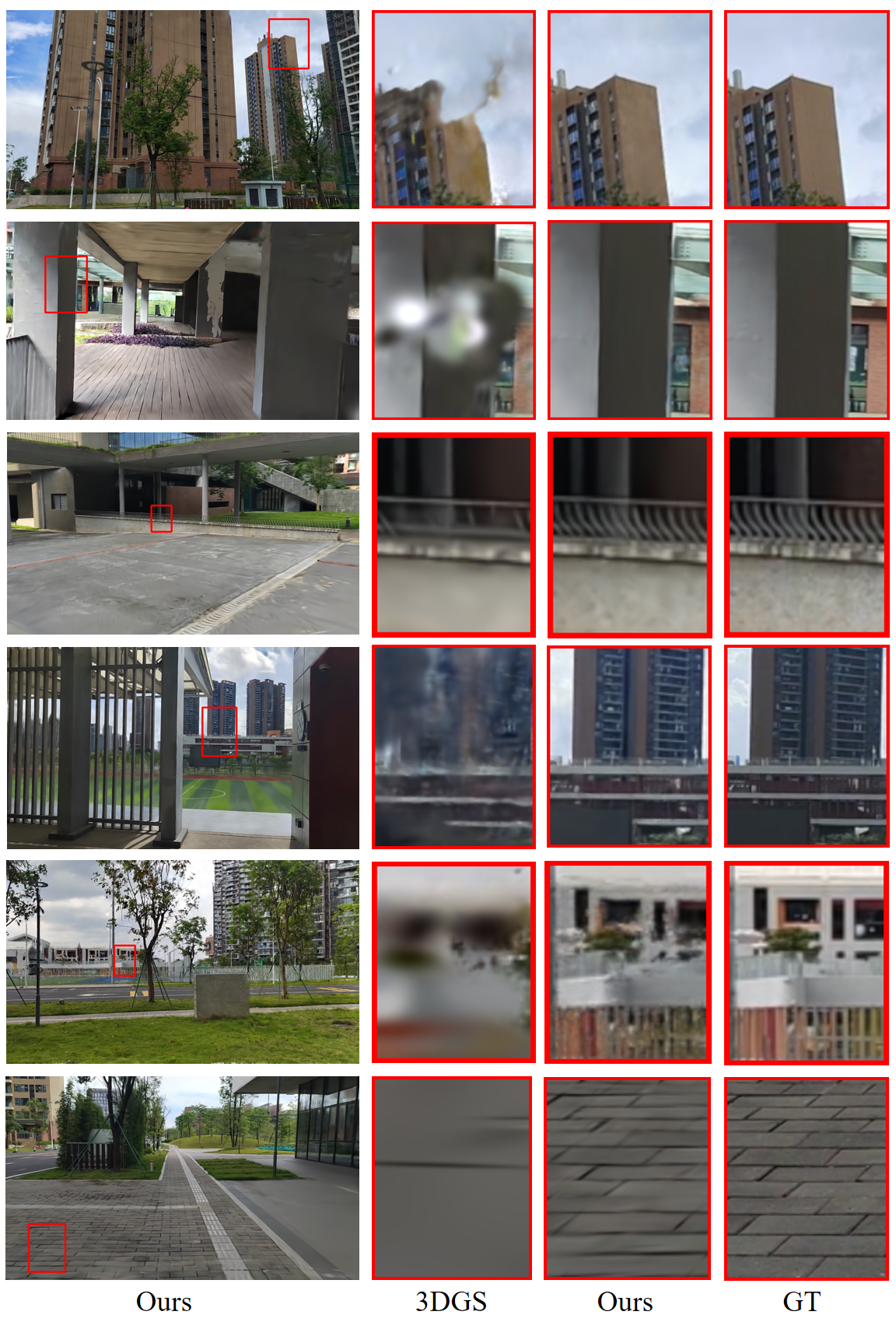} 
\caption{ Visual comparisons with 3DGS on our SCUTic dataset. 
}
\label{scutic_com}
\end{figure}

\subsection{Details of Our Dataset SCUTic}
We collect the dataset containing three large-scale free camera trajectories at the South China University of Technology International Campus, as shown in Figure \ref{camera_track}. We use a mobile phone to record videos, extract frames to obtain images, and then apply COLMAP to recover the poses of all images. The dataset includes three types of scenes: $general$, $indoor$, and $outdoor$, with 619 images in the $general$ scene, 605 images in the $indoor$ scene, and 647 images in the $outdoor$ scene. Our dataset includes rich texture details and numerous unbounded scenes, which can effectively validate the model’s representation capability. Figure \ref{scutic_com} compares the rendering effects of Toy-GS and 3DGS on the SCUTic dataset. Ours can render texture details and distant objects more accurately.

\subsection{Details of Adaptive Selection for Cameras}
In Section 3.1, we need to calculate the number of points visible to a camera $M_c$ and the number of visible points belonging to this area $M_{A_i}$ for adaptively selecting cameras.When calculating visibility points, in addition to filtering out points that project outside the image's range, occluded points that may weaken the effectiveness of adaptive spatial partitioning must also be considered. The solution is summarized as follows: First, we partition each image into multiple patches. The point cloud is then projected onto each image, where we preserve only the nearest projected point within each patch per image. This approach effectively eliminates occluded points. Finally, we use the k-means model trained by camera poses to compute the area of each point.

\end{document}